\documentclass[sigconf]{acmart}
\AtBeginDocument{%
  \providecommand\BibTeX{{%
    \normalfont B\kern-0.5em{\scshape i\kern-0.25em b}\kern-0.8em\TeX}}}

\usepackage{hyperref}       
\usepackage{url}            
\usepackage{booktabs}       

\usepackage{graphicx}
\usepackage{subcaption}
\usepackage{algpseudocode}
\usepackage{xspace}
\usepackage{tabu}
\usepackage{bm}
\usepackage{multirow}
\usepackage{enumitem}
\usepackage[ruled,vlined,linesnumbered]{algorithm2e}
\usepackage{color,soul}
\usepackage{wrapfig}
\usepackage{amsmath,multicol, mathrsfs}
\usepackage[bottom]{footmisc}
\usepackage{makecell}

\DeclareMathOperator*{\argmax}{arg\,max}

\newcommand{\ourSystem}{\textsc{SLiCE}}

\setcopyright{iw3c2w3}
\copyrightyear{2021}
\acmYear{2021}
\acmDOI{10.1145/3442381.3450060}

\acmConference[WWW '21]{Proceedings of the Web Conference 2021}{April 19--23, 2021}{Ljubljana, Slovenia}
\acmBooktitle{Proceedings of the Web Conference 2021 (WWW '21), April 19--23, 2021,
Ljubljana, Slovenia}
\acmPrice{}
\acmISBN{978-1-4503-8312-7/21/04}

\begin{document}

\title{Self-Supervised Learning of Contextual Embeddings for Link Prediction in Heterogeneous Networks}

\author{Ping Wang$^1$, Khushbu Agarwal$^2$, Colby Ham$^2$, Sutanay Choudhury$^2$, Chandan K. Reddy$^1$}
\affiliation{$^1$Department of Computer Science, Virginia Tech, Arlington, VA}
\affiliation{$^2$Pacific Northwest National Laboratory, Richland, WA}
\email{ping@vt.edu, {khushbu.agarwal, colby.ham, sutanay.choudhury}@pnnl.gov, reddy@cs.vt.edu}

\renewcommand{\shortauthors}{P. Wang, et al.}

\begin{abstract}
Representation learning methods for heterogeneous networks produce a low-dimensional vector embedding (that is typically fixed for all tasks) for each node. Many of the existing methods focus on obtaining a static vector representation for a node in a way that is agnostic to the downstream application where it is being used. In practice, however, downstream tasks such as link prediction require specific contextual information that can be extracted from the subgraphs related to the nodes provided as input to the task. To tackle this challenge, we develop $\ourSystem$, a framework for bridging static representation learning methods using global information from the entire graph with localized attention driven mechanisms to learn contextual node representations. We first pre-train our model in a self-supervised manner by introducing higher-order semantic associations and masking nodes, and then fine-tune our model for a specific link prediction task. Instead of training node representations by aggregating information from all semantic neighbors connected via metapaths, we automatically learn the composition of different metapaths that characterize the context for a specific task without the need for any pre-defined metapaths. $\ourSystem$ significantly outperforms both static and contextual embedding learning methods on several publicly available benchmark network datasets. We also demonstrate the interpretability, effectiveness of contextual learning, and the scalability of $\ourSystem$ through extensive evaluation. 
\end{abstract}




\maketitle

\section{Introduction}
The topic of representation learning for heterogeneous networks has gained a lot of attention in recent years \cite{dong2017metapath2vec, cen2019representation, yun2019graph, vashishth2019composition, wang2019heterogeneous, abu2018watch}, where a low-dimensional vector representation of each node in the graph is used for downstream applications such as link prediction \cite{zhang2018link, cen2019representation, abu2018watch} or multi-hop reasoning~\cite{hamilton2018embedding, das2016chains, zhang2018variational}. Many of the existing methods focus on obtaining a \textsl{static} vector representation per node that is agnostic to any specific context and is typically obtained by learning the importance of all of the node's immediate and multi-hop neighbors in the graph. However, we argue that nodes in a heterogeneous network exhibit a different behavior, based on different relation types and their participation in diverse network communities. Further, most downstream tasks such as link prediction are dependent on the specific contextual information related to the input nodes that can be extracted in the form of \textsl{task specific subgraphs}.

\begin{figure*}[htbp]
	\centering
	\begin{subfigure}[b]{0.3\textwidth}
		\includegraphics[width=\textwidth]{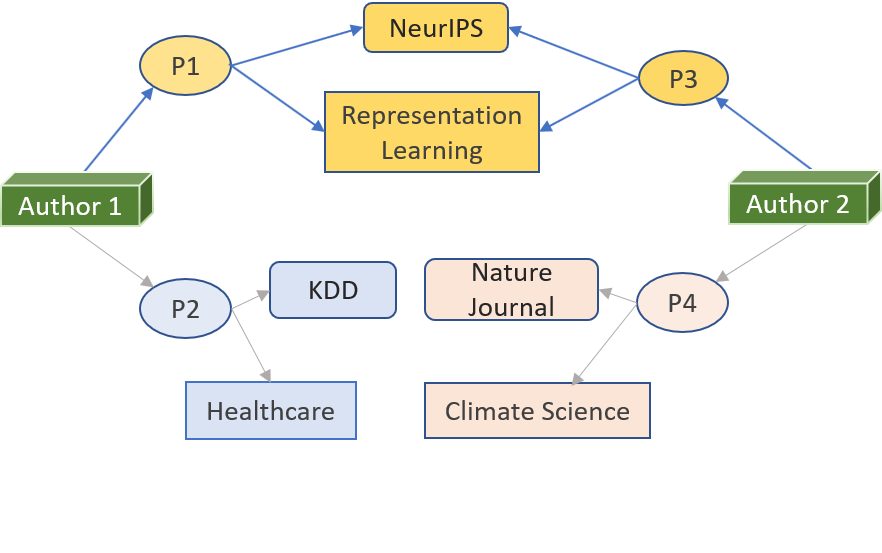}
		\caption{}\label{fig:1a}
	\end{subfigure}
	\hspace{2mm}
	\begin{subfigure}[b]{0.25\textwidth}
		\includegraphics[width=\textwidth]{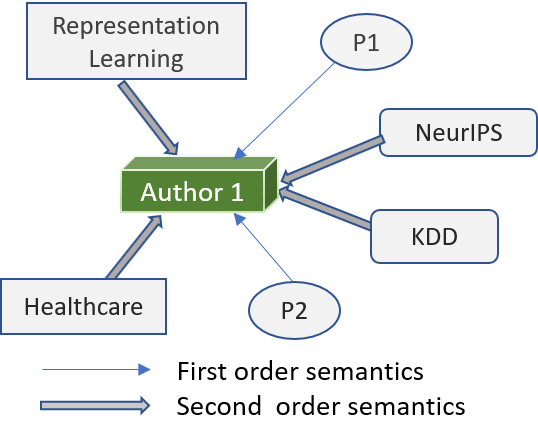}
		\caption{}\label{fig:1b}
	\end{subfigure}
	\hspace{2mm}
	\begin{subfigure}[b]{0.3\textwidth}
		\includegraphics[width=\textwidth]{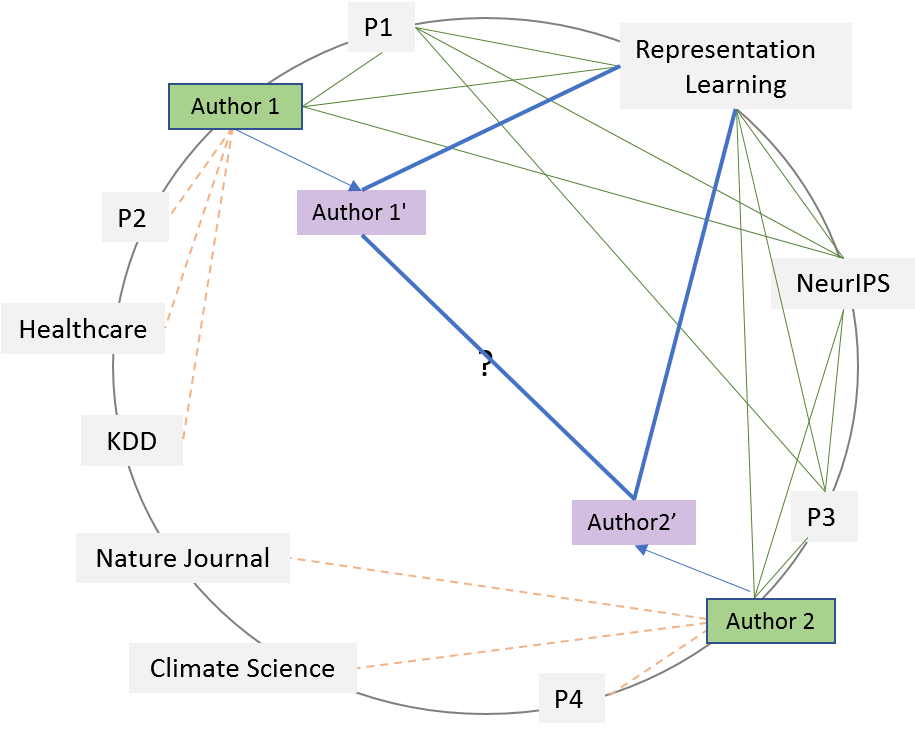}
		\caption{}\label{fig:1c}
	\end{subfigure}
	\vspace{-4.5mm}
	\caption{\small{Subgraph driven contextual learning in an academic network. (a) Author nodes publish on diverse topics (participate in diverse contexts) (b) State-of-the-art methods aggregate global semantics for authors based on all published papers (c) Our approach uses context subgraph between authors to contextualize their node embeddings during link prediction.}}
	\vspace{-4mm}
	\label{fig:motivation}
\end{figure*}

Incorporation of contextual learning has led to major breakthroughs in the natural language processing community \cite{peters2018deep, devlin2018bert}, in which the same word is associated with different concepts depending on the context of the surrounding words. A similar phenomenon can be exploited in graph structured data and it becomes particularly pronounced in heterogeneous networks where the addition of relation types as well as node and relation attributes leads to increased diversity in a node's contexts. Figure \ref{fig:motivation} provides an illustration of this problem for an academic network. Given two authors who publish in diverse communities, we posit that the task of predicting link $(Author_1, co-author, Author_2)$ would perform better if their node representation is reflective of the common publication topics and venues, i.e., Representation Learning and NeurIPS. This is in contrast to existing methods where author embeddings would reflect information aggregation from all of their publications, including the publications in healthcare and climate science which are not part of the common context.

Contextual learning of node representations in network data has recently gained attention with different notions of context emerging (see Table \ref{tab:related_work}).  In homogeneous networks, communities provide a natural definition of a node's participation in different contexts referred to as facets or aspects \cite{yang2018multi, epasto2019single, liu2019single, wang2019mcne, park2020unsupervised}. Given a task such as link prediction, inferring the cluster-driven connectivity between the nodes has been the primary basis for these approaches. However, accounting for higher-order effects over diverse meta-paths (defined as paths connected via heterogeneous relations) is demonstrated to be essential in representation learning and link prediction in heterogeneous networks \cite{cen2019representation, yun2019graph, wang2019heterogeneous, hu2020heterogeneous}. Therefore, contextual learning methods that primarily rely on the well-defined notion of graph clustering will be limited in their effectiveness for heterogeneous networks where modeling semantic association (via meta-paths or meta-graphs) is at least equal or more important than community structure for link prediction.

In this paper, we seek to make a fundamental advancement over these categories that aim to contextualize a node's representation with regards to either a cluster membership or association with meta-paths or meta-graphs. We believe that the definition of a context needs to be expanded to subgraphs (comprising heterogeneous relations) that are task-specific and learn node representations that represent the collective heterogeneous context. With such a design, a node's embedding will be dynamically changing based on its participation in one input subgraph to another. Our experiments indicate that this approach has a strong merit with link prediction performance, thus improving it by 10\%-25\% over many state-of-the-art approaches.

We propose shifting the node representation learning from a node's perspective to a subgraph point of view. Instead, of focusing on ``what is the best representation for a node $v$", we seek to answer ``what are the best collective node representations for a given subgraph $g_c$" and ``how such representations can be potentially useful in a downstream application?" Our proposed framework $\ourSystem$ (which is an acronym for Self-supervised LearnIng of Contextual Embeddings), accomplishes this by bridging static representation learning methods using global information from the entire graph with \textit{localized attention driven mechanisms to learn contextual node representations in heterogeneous networks.}  While bridging global and local information is a common approach for many algorithms, the primary novelty of $\ourSystem$ lies in learning an operator for contextual translation by learning higher-order interactions through self-supervised learning. \vspace{0.04in}

\begin{figure*}[!tp]
  \centering
  \vspace{-4mm}
  \includegraphics[width=0.9\textwidth]{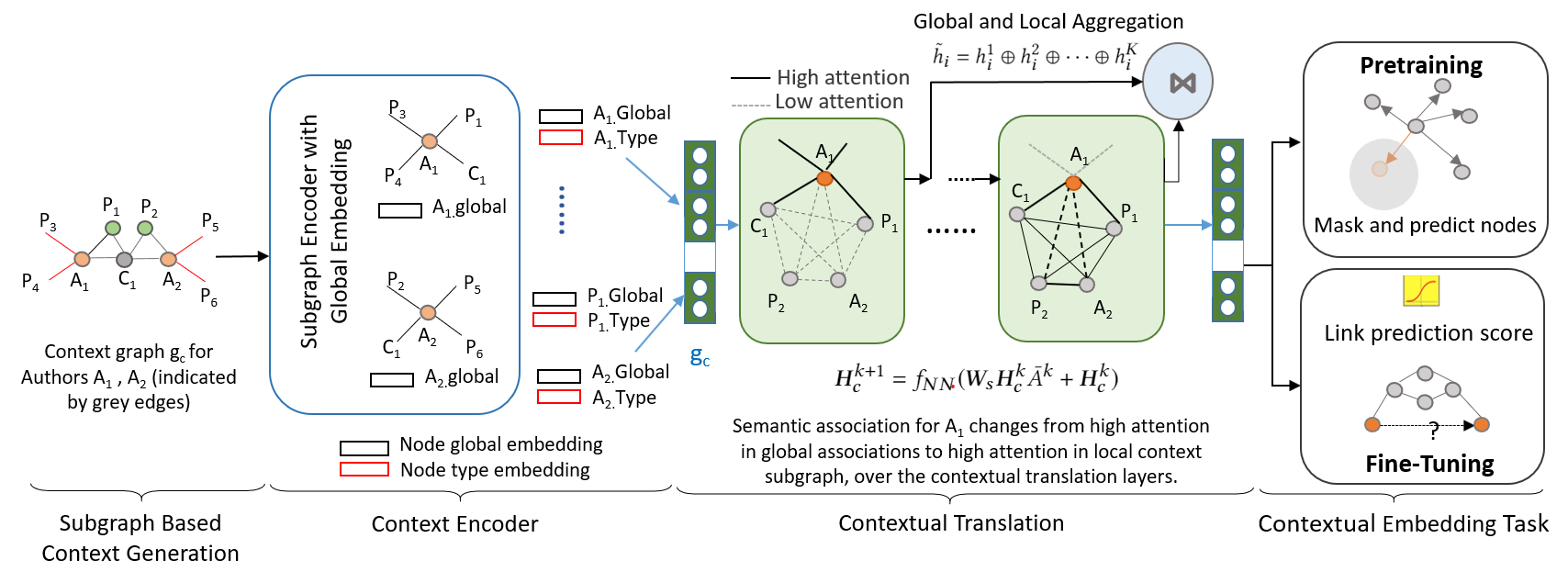}
  \vspace{-4mm}
  \caption{\small{Overview of $\ourSystem$ architecture. Subgraph context is initialized using  global features for each node. Each layer in $\ourSystem$ shifts the embedding of all nodes in $g_c$ to emphasize the local dependencies in the contextual subgraph. The final embeddings for nodes in context subgraphs are determined as a function of output from last $i$ layers to combine global with local contextual semantics for each node.}}
  \label{fig:neural_architecture}
  \vspace{-4mm}
\end{figure*}

\noindent \textbf{Contextualized Representations:} Building on the concept of \textsl{translation-based embedding models} \cite{bordes2013translating}, given a node $u$ and its embedding $h_u$ computed using a global representation method, we formulate graph-based learning of contextual embeddings as performing a vector-space translation $\Delta_u$ (informally referred to as \textsl{shifting process}) such that $h_u + \Delta_u \approx \Tilde{h}_u$, where $\Tilde{h}_u$ is the contextualized representation of $u$. The key idea behind $\ourSystem$ is to learn the translation $\Delta_u$ where $u \in V(g_c)$. Figure \ref{fig:motivation}(c) shows an illustration of this concept where the embedding of both Author1 and Author2 are shifted using the common subgraph with (Paper P1, Representation learning , NeurIPS, Paper P3) as context. We achieve this contextualization as follows: We first learn the {higher-order semantic association} (HSA) between nodes in a context subgraph. We do not assume any prior knowledge about important metapaths, and $\ourSystem$ learns important task specific subgraph structures during training (see section 4.3). More specifically, we first develop a \textsl{self-supervised learning} approach that pre-trains a model to learn a HSA matrix on a context-by-context basis. We then fine-tune the model in a task-specific manner, where given a context subgraph $g_c$ as input, we encode the subgraph with global features and then transform that initial representation via a HSA-based non-linear transformation to produce contextual embeddings (see Figure \ref{fig:neural_architecture}). \vspace{0.04in}

\noindent \textbf{Our Contributions:} The main contributions of our work are:

\begin{itemize}[leftmargin=*]
\item Propose \textit{contextual embedding learning for graphs from single relation context to arbitrary subgraphs}.  

\item Introduce a \textit{novel self-supervised learning approach to learn higher-order semantic associations between nodes} by simultaneously capturing the global and local factors that characterize a context subgraph. 

\item Show that $\ourSystem$ significantly outperforms existing static and contextual embedding learning methods using standard evaluation metrics for the task of link prediction. 

\item Demonstrate the interpretability, effectiveness of contextual translation, and the scalability of $\ourSystem$ through an extensive set of experiments and contribution of a new benchmark dataset.
\end{itemize}

The rest of this paper is organized as follows. Section~\ref{sec:related} provides an overview of related work about network embedding learning and differentiates our work from other existing work. In Section~\ref{sec:model}, we introduce the problem formulation and present the proposed  $\ourSystem$ model. We describe the details of experimental analysis and show the comparison of our model with the state-of-the-art methods in Section~\ref{sec:exp}. Finally, Section~\ref{sec:conclusion} conclude the discussion of the paper.

\begin{table*}[!tp]
\caption{Comparison of representative approaches for learning heterogeneous network (HN) embeddings proposed in the recent literature from contextual learning perspective. Other abbreviations used: graph convolutional network (GCN), graph neural network (GNN), random walk (RW), skip-gram (SG). N/A stands for ``Not Applicable''.}
\vspace{-2mm}
\resizebox{\linewidth}{!}{
\begin{tabular}{|l|c|c|c|c|c|}
\hline
\bf Method &\bf \makecell{Multi-Embeddings\\ per Node} &\bf Context Scope &\bf HN support &\bf Learning Approach &\bf \makecell{Automated Learning\\ of Meta-Paths/Graphs} \\
\hline
HetGNN \cite{zhang2019heterogeneous}, HetGAT \cite{wang2019heterogeneous}  & N & N/A & Y & GNN & N  \\\hline
GTN \cite{yun2019graph}, HGT \cite{hu2020heterogeneous} & N & N/A & Y & Transformer & Y  \\\hline
GAN \cite{abu2018watch}, RGCN \cite{schlichtkrull2018modeling} & N & N/A & Y & GCN & N  \\\hline
Polysemy \cite{liu2019single}, MCNE \cite{wang2019mcne} & Y & Per aspect  & N & Extends SG/GCN, GNN  & N  \\\hline
GATNE \cite{cen2019representation}, CompGCN \cite{vashishth2019composition} & Y & Per relation & Y & HN-SG, GCN & N/A  \\\hline
SPLITTER \cite{epasto2019single}, asp2vec \cite{park2020unsupervised} & Y & Per aspect & Y & Extends RW-SG & N  \\\hline
SLiCE (proposed) & Y & Per subgraph & Y & Self-supervision & Y  \\\hline
\end{tabular}
}
\label{tab:related_work}
\vspace{-3mm}
\end{table*}

\section{Related Work}
\label{sec:related}
We begin with an overview of the state-of-the-art methods for representation learning in heterogeneous network and then follow with a discussion on the nascent area of contextual representation learning. Table \ref{tab:related_work} provides a summarized view of this discussion. \vspace{0.04in}

\noindent \textbf{Node Representation Learning:} Earlier representation learning algorithms for networks can be broadly categorized into two groups based on their usage of matrix factorization versus random walks or skip-gram-based methods.  Given a graph $G$, matrix factorization based methods \cite{cao2015grarep} seek to learn a representation $\Gamma$ that minimizes a loss function of the form $||\Gamma^T\Gamma - P_V||^2$, where $P_V$ is a matrix containing pairwise proximity measures for $V(G)$.  Random walk based methods such as DeepWalk \cite{perozzi2014deepwalk} and node2vec \cite{grover2016node2vec} try to learn representations that approximately minimize a cross-entry loss function of the form $\sum_{v_i, v_j \in V(G)} -log(p_L(v_j | v_i))$, where $p_L(v_j | v_i)$ is the probability of visiting a node $v_j$ on a random walk of length $L$ starting from node $v_i$.  Node2vec based approach has been further extended to incorporate multi-relational properties of networks by constraining random walks to ones conforming to specific metapaths \cite{dong2017metapath2vec, chen2018pme}. Recent efforts \cite{qiu2018network} seek to unify the first two categories by demonstrating the equivalence of \cite{perozzi2014deepwalk} and \cite{grover2016node2vec}-like methods to matrix factorization approaches. \vspace{0.04in}

\noindent \textbf{Attention-based Methods:} A newer category represents graph neural networks and their variants \cite{kipf2016semi, schlichtkrull2018modeling}. Attention mechanisms that learn a distribution for aggregating information from a node's immediate neighbors is investigated in \cite{velivckovic2017graph}.  Aggregation of attention from \textsl{semantic neighbors}, or nodes that are connected via multi-hop metapaths have been exhaustively investigated over the past few years and can be grouped by the underlying neural architectures such as graph convolutional networks \cite{vashishth2019composition}, graph neural network \cite{zhang2019heterogeneous, wang2019heterogeneous}, and graph transformers \cite{yun2019graph, hu2020heterogeneous}. Extending the above methods from meta-paths to meta-graphs \cite{he2019hetespaceywalk, zhang2020mg2vec} as a basis for sampling and learning has emerged as a new direction as well.\vspace{0.04in}

\noindent \textbf{Contextual Representation Learning:} Works such as \cite{yang2018multi, epasto2019single, liu2019single, wang2019mcne, sun2019vgraph, bandyopadhyay2020outlier} study the ``multi-aspect'' effect. Typically, ``aspect'' is defined as a node's participation in a community or cluster in the graph or even being an outlier, and these methods produce a node embedding by accounting for it's membership in different clusters.  However, most of these methods are studied in detail for homogeneous networks. More recently, this line of work has evolved towards addressing finer issues such as inferring the context, addressing limitations with offline clustering, producing different vectors depending on the context as well as extension towards heterogeneous networks \cite{ma2019disentangled, park2020unsupervised}.

Beyond these works, a number of newer approaches and objectives have also emerged. The authors of \cite{abu2018watch} compute a node's representation by learning the attention distribution over a graph walk context where it occurs. The work presented in \cite{cen2019representation} is a metapath-constrained random walk based method that \textsl{contextualizes} node representations per relation. It combines a base embedding derived from the global structure (similar to above methods) with a relation-specific component learnt from the metapaths. In a similar vein, \cite{vashishth2019composition} provides operators to adapt a node's embedding based on the associated relational context.\vspace{0.04in}

\noindent \textbf{Key Distinctions of \ourSystem:} To summarize, the key differences between $\ourSystem$ and existing works are as follows:
(i) \textit{Our contextualization objective} is formulated on the basis of a subgraph that distinguishes $\ourSystem$ from \cite{cen2019representation} and \cite{vashishth2019composition}. While subgraph-based representation learning objectives have been superficially investigated in the literature \cite{zhang2019hyper}, they do not focus on generating contextual embeddings. (ii) From a \textit{modeling perspective}, our self-supervised approach is distinct from both the metapath-based learning approaches outlined in the \textsl{attention-based methods section} (we learn important metapaths automatically without requiring any user supervision) as well as the clustering-centric multi-aspect approaches discussed in the \textsl{contextual representation learning} category.
\vspace{-2mm}
\section{The Proposed Framework}
\label{sec:model}

\subsection{Problem Formulation}
Before presenting our overall framework, we first briefly provide the formal definitions and notations that are required to comprehend the proposed approach.

\noindent \textsc{\textbf{Definition: (Heterogeneous Graph)}}. We represent a heterogeneous graph as a 6-tuple $G = (V, E, \Sigma_V, \Sigma_E, \lambda_v, \lambda_e)$ 
where, $V$ (alternatively referred to as $V(G)$) is the set of nodes and $E$ (or $E(G)$) denotes the set of edges between the nodes. 
$\lambda_v$ and $\lambda_e$ are functions mapping the node (or edge) to its node (or edge) type $\Sigma_V$ and $\Sigma_E$, respectively. \vspace{0.04in}

\noindent \textsc{\textbf{Definition: (Context Subgraph)}}. Given a heterogeneous graph $G$, the context of a node $v$ or node-pair $(u, v)$ in $G$ can be represented as the subgraph $g_c$ that includes a set of nodes selected with certain criteria (e.g., $k$-hop neighbors for $v$ or $k$-hop neighbors connecting $(u, v)$) along with their related edges. The context of the node or node-pair can be represented as $g_c(v)$ and $g_c(u,v)$. \vspace{0.04in}

\noindent \textsc{\textbf{Problem definition}}.
Given a heterogeneous graph $G$, a subgraph $g_c$ and the link prediction task $T$, compute a function $f(G, g_c, T)$ that maps each node $v_i$ in the set of vertices in $g_c$, denoted as $V(g_c)$, to a real-valued embedding vector $h_i$ in a low-dimensional space $d$ such that $h_i \in \mathbb{R}^d$. We also require that $S$, a scoring function serving as a proxy for the link prediction task, satisfies the following: $S(v_i, v_j) \geq \delta$ for a positive edge $e=(v_i, v_j)$ in graph $G$ and $S(v_i, v_j) < \delta$ if $e$ is a negative edge, where $\delta$ is a threshold.  \vspace{0.04in}

\noindent \textsc{\textbf{Overview of \ourSystem}}. In this paper, the proposed $\ourSystem$ framework mainly consists of the following four components: 
(1) \textit{\textbf{Contextual Subgraph Generation and Encoding}}: generating a collection of context subgraphs and transforming them into a vector representation.
(2)  \textit{\textbf{Contextual Translation}}: for nodes in a context subgraph, translating the global embeddings, which consider various heterogeneous attributes about nodes, relations and graph structure, to contextual embeddings based on the specific local context.
(3) \textit{\textbf{Model Pre-training}}: learning higher order relations with the self-supervised contextualized node prediction task. 
(4) \textit{\textbf{Model Fine-tuning}}: the model is then tuned by the supervised link prediction task with more fine-grained contexts for node pairs. Figure~\ref{fig:neural_architecture} shows the framework of the proposed $\ourSystem$ model. In the following sections, we will introduce more details layer-by-layer. 

\vspace{-2mm}
\subsection{Context Subgraphs: Generation and Representation}
\label{sec:context_generation}
\subsubsection{Context Subgraph Generation}
In this work, we generate the pre-training subgraphs $G_C$ for each node in the graph using a random walk strategy. Masking and predicting nodes in the random walks helps $\ourSystem$ learn the global connectivity patterns in $G$ during the pre-training phase. In fine tuning phase, the context subgraphs for link prediction are generated using following approaches: (1) \textit{Shortest Path} strategy considers the shortest path between two nodes as the context. (2) \textit{Random} strategy, on the other hand, generates contexts following the random walks between two nodes, limited to a pre-defined maximum number of hops.
Note that the context generation strategies are generic and can be applied for generating contexts in many downstream tasks such as link prediction \cite{zhang2018link}, knowledge base completion \cite{socher2013reasoning} or multi-hop reasoning \cite{das2016chains, hamilton2018embedding}.

In our experiments, context subgraphs are generated for each node $v$ in the graph during pre-training and for each node-pair during fine-tuning.
Each generated subgraph $g_c\in G_c$ is encoded as a set of nodes denoted by $g_c=(v_1, v_2, \cdots, v_{|V_c|})$, where $|V_c|$ represents the number of nodes in $g_c$.  Different from the sequential orders enforced on graph sampled using pre-defined metapaths, the order of nodes in this set is not important. Therefore, our context subgraphs are not limited to paths, and can handle tree or star-shaped subgraphs.

\vspace{-2mm}
\subsubsection{Context Subgraph Encoder}
We first represent each node $v_i$ in the context subgraph as a low-dimensional vector representation by $h_i=\bm{W}_e f_{attr}(v_i)$, where $f_{attr}(.)$ is a function that returns a stacked vector containing the structure-based embedding of $v_i$ and the embedding of its attributes.  $\bm{W}_e$ is the learnable embedding matrix. We represent the input node embeddings in $g_c$ as $\bm{H}_c=(h_1, h_2, \cdots, h_{|V_c|})$. It is flexible to incorporate the node and relation attributes (if available) for attributed networks \cite{cen2019representation} in the low-dimensional representations or initialize them with the output embeddings learnt from other global feature generation approaches that capture the multi-relational graph structure~\cite{grover2016node2vec, dong2017metapath2vec, wang2019heterogeneous, yun2019graph, vashishth2019composition}.

There are multiple approaches for generating global node features in heterogeneous networks (see ``related work"). 
Our experiments show that the node embeddings obtained from random walk based skip-gram methods (RW-SG) produces competitive performance for link prediction tasks. Therefore, in the proposed $\ourSystem$ model, we mainly consider the pre-trained node representation vectors from node2vec for initialization of the node features. 

\vspace{-2mm}
\subsection{Contextual Translation}
Given a set of nodes $V_c$ in a context subgraph $g_c$ and their global input embeddings $\bm{H}_c \in \mathbb{R}^{d \times |V_c|}$, the primary goal of contextual learning is to \textit{translate (or shift)} the global embeddings
in the vector space towards their new positions that indicate the most representative roles of nodes in
the structure of $g_c$. We consider this mechanism as a transformation layer and the model can include multiple such layers according to the higher-order relations contained in the graph. Before introducing the details of this contextual translation mechanism, we first provide the definition of the \textit{semantic association matrix}, which serves as the primary indicator about the translation of embeddings according to specific contexts. \vspace{0.04in}

\noindent \textsc{\textbf{Definition: {(Semantic Association Matrix)}}}. A semantic association matrix, denoted as $\bar{A}$, is an \textit{asymmetric weight matrix} that indicates the high-order relational dependencies between nodes in the context subgraph $g_c$. 


Note that the semantic association matrix will be asymmetric since the influences of two nodes on one another in a context subgraph tend to be different. The adjacency matrix of the context subgraph, denoted by $A_{gc}$, can be considered as a trivial candidate for $\bar{A}$, which includes the local relational information of context subgraph $g_c$.
However, the goal of contextual embedding learning is to translate the global embeddings using the contextual information contained in
the specific context $g_c$ while keeping the nodes' connectivity through the global graph. Hence, instead of setting it to $A_{gc}$, we contextually learn the semantic associations, or more specifically the weights of the matrix $\bar{A}^k$ in each translation layer $k$ by incorporating the connectivity between nodes through both local context subgraph $g_c$ and global graph $G$. \vspace{0.04in}

\noindent \textbf{Implementation of Contextual Translation:}
In the translation layer $k+1$, the semantic association matrix $\bar{A}^k \in \mathbb{R}^{|V_c| \times |V_c|}$ is updated by the transformation operation defined in Eq.~(\ref{eqn:embed_shift}). It is accomplished by performing message passing across all nodes in context subgraph $g_c$ and updating the node embeddings $\bm{H}_c^k=(h_1^k, h_2^k, \cdots, h_{|V_c|}^k)$ to be $\bm{H}_c^{k+1}$.
\begin{equation}
    \bm{H}^{k+1}_c = f_{NN}(\bm{W}_s \bm{H}^k_c \bar{A}^k  + \bm{H}^k_c)
    \label{eqn:embed_shift}
\end{equation}
where $f_{NN}$ is a non-linear function and the transformation matrix $\bm{W}_s \in \mathbb{R}^{d \times d}$ is the learnable parameter. The residual connection \cite{he2016deep} is applied to preserve the contextual embeddings in the previous step. This allows us to still maintain the global relations by passing the original global embeddings through layers while learning contextual embeddings.  
Given two nodes $v_i$ and $v_j$ in context subgraph $g_c$, the corresponding entry $\bar{A}^k_{ij}$ in semantic association matrix can be computed using the multi-head (with $N_h$ heads) attention mechanism~\cite{vaswani2017attention} in order to capture relational dependencies under different subspaces. For each head, we calculate $\bar{A}^k_{ij}$ as follows: 
\begin{equation}
\bar{A}^k_{ij} = \frac{exp\left((\bm{W}_1 h^k_i)^T (\bm{W}_2 h^k_j)\right)}{\sum_{t=1}^{|V_c|} exp\left((\bm{W}_1 h^k_i)^T (\bm{W}_2 h^k_t)\right)}
\label{eqn:a_matrix}
\end{equation}
where the transformation matrix $\bm{W}_1$ and $\bm{W}_2$ are learnable parameters. 
It should be noted that, different from the aggregation procedure performed across all nodes in the general graph G, the proposed translation operation is only performed within the local context subgraph $g_c$. 
The updated embeddings after applying the translation operation according to context $g_c$ indicate the most representative roles of each node in the specific local context neighborhood. 
In order to capture the higher-order association relations within the context, we apply multiple layers of the transformation operation in Eq.~(\ref{eqn:embed_shift}) by stacking $K$ layers as shown in Figure~\ref{fig:neural_architecture}, where $K$ is the largest diameter of the subgraphs sampled in context generation process.

\begin{algorithm}[tp]
	\SetAlgoLined
	\SetKw{Initialize}{Initialize}
	\textbf{Require:} Graph $G$ with nodes set $V$ and edges set $E$, embedding size $d$, context subgraph size $m$, No. of translation layers $K$.\\
	Pre-training dataset $G_c^{pre}\leftarrow \emptyset$\\
	\For{\textbf{each} $v \in V$}{
    $g^v_c=$GetContext($G$, $v$, $m$)\ \  $\rhd$ Generate\ context\ subgraph\\
    $g^v_c=Encode(g_c^v)$\ \ \ \ \ \ \ \ \ \ \ $\rhd$ Encode\ context\ as\ a\ sequence \\
    \emph{$v_m=Random(g_c^v)$\ \ \ \ \  \ $\rhd$ Mask\ a\ node\ in\ $g^v_c$ for\ prediction}\\
    $G_c^{pre}.Append(g^v_c, v_m)$
    }
    $\bm{H} \leftarrow EmbedFunction(G, d)$\ \ \ \ \ \ \ \ \ $\rhd$ Learn\ global\ embeddings\\
    Initialize node embeddings as $\bm{H}^0=\bm{H}$.\\
    \While{not converged}{
	\For{$g_c^v$ \KwTo $G_c^{pre}$}{
	\For{$k= 1$ \KwTo $K$}{
	$\bm{H}^{k+1}_c = f_{NN}(\bm{W}_s \bm{H}^k_c \bar{A}^k + \bm{H}^k_c)$ with $\bar{A}$ calculated by Eq.~(\ref{eqn:a_matrix})\\
	}
	\emph{Contextual embeddings $\Tilde{\bm{H}}_c=\bm{H}_c^1\oplus \bm{H}_c^2\oplus \dots \oplus \bm{H}_c^K$}\\
	
	\emph{Update parameters with the contextualized node prediction task using the objective function in Eq.~(\ref{eqn:learning_obj_pre_training}).}
	}
	}
	\caption{Self-supervised Pre-training in $\ourSystem$.}
	\label{alg:pre}
\end{algorithm}

By applying multiple translation layers, we are able to obtain multiple embeddings for each node in the context subgraph. In order to collectively consider different embeddings in the downstream tasks, we aggregate the node embeddings learnt from different layers $\{h_i^k\}_{k=1,...,K}$ as the contextual embedding $\Tilde{h}_i$ for each node as follows.
\begin{equation}
    \Tilde{h}_i = h_i^1\oplus h_i^2\oplus \dots \oplus h_i^K
    \label{eqn:concat}
\end{equation}
Given a context subgraph $g_c$, the obtained contextual embedding vectors $\{\Tilde{h}_i\}_{i=1,2,...,|V_c|}$ can be fed into the prediction tasks.  
In pre-training step, a linear projection function is applied on the contextual embeddings to predict the probability of masked nodes. For fine-tuning step, we apply a single layer feed-forward network with softmax activation function for binary link prediction.

\subsection{Model Training Objectives}
\subsubsection{Self-supervised Contextual Node Prediction}
Our model pre-training is performed by training the self-supervised contextualized node prediction task. More specifically, for each node in $G$, we generate the node context $g_c$ with diameter (defined as the largest shortest pair between any pair of nodes) using the aforementioned context generation methods and randomly mask a node for prediction based on the context subgraph. The graph structure is left unperturbed by the masking procedure. Therefore, the pre-training is learnt by maximizing the probability of observing this masked node $v_{m}$ based on the context $g_c$ in the following form. \begin{equation}
\theta =  \argmax_\theta \prod_{g_c \in G_C} \prod_{v_{m} \in g_{c}} p(v_{m} | g_c, \theta)
\label{eqn:learning_obj_pre_training}
\end{equation}
where $\theta$ represents the set of model parameters. The procedure for pre-training is given in Algorithm~\ref{alg:pre}. 
{In this algorithm, lines 2-8 generate context subgraphs for nodes in the graph and further applies a random masking strategy to process the data for pre-training. Lines 9-10 learn the pre-trained global node features and initialize them as the node embeddings in $\ourSystem$. In lines 13-15, we apply the contextual translation layers on the context subgraphs, aggregate the output of different layers as the contextual node embeddings in line 16 and update the model parameters with contextual node prediction task.}
In a departure from traditional skip-gram methods~\cite{park2020unsupervised} that predicts a node from the path prefix that precedes it in a random walk, our random masking strategy forces the model to learn higher-order relationships between nodes that are arbitrarily connected by variable length paths with diverse relational patterns.  

\subsubsection{Fine-tuning with Supervised Link Prediction} 
The $\ourSystem$ model is further fine-tuned on the \textit{contextualized link prediction task by generating multiple fine-grained contexts for each specific node-pair} that is under consideration for link prediction. Based on the predicted scores, this stage is trained by maximizing the probability of observing a positive edge ($e_p$) given context~($g_{cp}$), while also learning to assign low probability to negatively sampled edges ($e_n$) and their associated contexts~($g_{cn}$). The overall objective is obtained by summing over the training data subsets with positive edges~($D_p$) and negative edges ($D_n$). 
Algorithm~\ref{alg:fine} shows the process of fine-tuning step. {In this algorithm, lines 2-7 generate context subgraphs of the node-pairs for link prediction task and process the data for fine-tuning in the same manner described in pre-training. Lines 8-18 perform the fine-tuning with link prediction task.}
\begin{equation}
\mathcal{L}= \sum_{\left(e_{p}, g_{cp}\right) \in D_{p}}  \log(P(e_{p} | g_{cp}, \theta)) +   \sum_{\left(e_{n}, g_{cn}\right) \in D_n} \log(1 - P(e_n | g_{cn}, \theta)) 
\label{eqn:learning_obj_fine_tuning}
\end{equation}
We compute the probability of the edge between two nodes $e=(v_i, v_j)$ as the similarity score $S(v_i, v_j) = \sigma(\Tilde{h}_i^T\cdot \Tilde{h}_j)$~\cite{abu2018watch}, where $\Tilde{h}_i$ and $\Tilde{h}_j$ are contextual embeddings of $v_i$ and $v_j$ learnt based on a context subgraph, respectively. $\sigma(\cdot)$ represents sigmoid function. 

\begin{algorithm}[tp]
    \SetAlgoLined
    \SetKw{Initialize}{Initialize}
    \textbf{Require:} Graph $G$ with nodes set $V$ and edges set $E$, list of edges $E_{lp}$ for link prediction, global embeddings $\bm{H}$, pre-trained model parameters $\theta$, context subgraph size $m$, No. of translation layers $K$.\\
    
    Fine-tuning dataset $G_c^{fine}\leftarrow \emptyset$\\
	\For{\textbf{each} $e \in E_{lp}$}{
    $g^e_c=$GetContext($G$, $e$, $m$)\ \  $\rhd$ Generate\ context\ subgraph\\
    $g^e_c=Encode(g_c^e)$\ \ \ \ \ \ \ \ \ \ \ $\rhd$ Encode\ context\ as\ a\ sequence \\
    $G_c^{fine}.Append(g^e_c)$
    }
    Initialize node embeddings as $\bm{H}^0=\bm{H}$.\\
    Set model parameters to pre-trained parameters $\theta$ in Algorithm~\ref{alg:pre}.\\
    \While{not converged}{
	\For{$g_c^e$ \KwTo $G_c^{fine}$}{
	\For{$k= 1$ \KwTo $K$}{
	$\bm{H}^{k+1}_c = f_{NN}(\bm{W}_s \bm{H}^k_c \bar{A}^k + \bm{H}^k_c)$ with $\bar{A}$ calculated by Eq.~(\ref{eqn:a_matrix})\\
	}
	\emph{Contextual embeddings $\Tilde{\bm{H}}_c=\bm{H}_c^1\oplus \bm{H}_c^2\oplus \dots \oplus \bm{H}_c^K$}\\
	
	\emph{Update fine-tuning parameters using  Eq.~(\ref{eqn:learning_obj_fine_tuning}).}
	}
	}
    \caption{Fine-tuning in $\ourSystem$ with link prediction.}
    \label{alg:fine}
\end{algorithm}

\subsection{Complexity Analysis}
\label{sec:complexity}
We assume that $N_{cpn}$ denotes the number of context subgraphs generated for each node, $N_{mc}$ represents the maximum number of nodes in any context subgraph, and $|V|$ represents the number of nodes in the input graph $G$. 
Then, the total number of context subgraphs considered in pre-training stage can be estimated as $|V|*N_{cpn}$ and the cost of iterating over all these subgraphs through multiple epochs will be $O(|V|*N_{cpn})$. Since the generated context subgraphs need to provide us with a good approximation of the total number of edges in the entire graph, we approximate the total cost as $O(|V|*N_{cpn}) \approx O(N_E)$, where $N_E$ is the number of edges in the training dataset. It can also be represented as $N_E = \alpha_T |E|$, where $|E|$ is the total number of edges in graph $G$ and $\alpha_T$ represents the ratio of training split. 
The cost for each  contextual translation layer in $\ourSystem$ model is $O(N^2_{mc})$ since the dot product for calculating node similarity is the dominant computation and is quadratic to the number of nodes in the context subgraph. In this case, the total training complexity will be $O(|E|N^2_{mc})$. The maximum number of nodes $N_{mc}$ in context subgraphs is relatively small and it can be considered as a constant that does not depend on the size of the input graph. Therefore, the training complexity of $\ourSystem$ is approximately linear to the number of edges in the input graph.

\subsection{Implementation Details} 
\label{sec: slice-implementation}
{The proposed $\ourSystem$ model is implemented using PyTorch 1.3} \cite{paszke2017automatic}. The dimension of contextual node embeddings is set to 128 in \ourSystem.
We used a skip-gram based random walk approach to encode context subgraphs with global node features. 
Both pre-training and fine-tuning steps in $\ourSystem$ are trained for 10 epochs with a batch size of 128 using the cross-entropy loss function. The model parameters are trained with ADAM optimizer~\cite{kingma2014adam} with a learning rate of 0.0001 and 0.001 for pre-training and fine-tuning steps, respectively. The best model parameters were selected based on the development set. 
Both the number of contextual translation layers and number of self-attention heads are set to 4.
We generate context subgraphs by performing random walks between node pairs by setting the maximum number of nodes in context subgraphs and the number of contexts generated for each node to be \{6, 12\} and \{1, 5, 10\}, respectively and report the best performance. In the fine-tuning stage, the subgraph with the largest prediction score are selected as the best context subgraph for each node-pair. 
The implementation of the $\ourSystem$ model is made publicly available at this website\footnote{https://github.com/pnnl/SLICE}.

\section{Experiments}
\label{sec:exp}

{In this section, we address following research questions (RQs) through experimental analysis}: 

\begin{enumerate}[leftmargin=*]
    \item \textbf{RQ1}: Does subgraph-based contextual learning improve the performance of downstream tasks such as link prediction? 
    \item \textbf{RQ2}: Can we interpret $\ourSystem$'s performance using semantic features of context subgraphs?
    \item \textbf{RQ3}: Where does contextualization help in graphs? How do we quantify the effect of the embedding shift from static to subgraph-based contextual learning?
    \item \textbf{RQ4}: What is the impact of different parameters and components (including pre-trained global features and fine-tuning procedure) on the performance of $\ourSystem$? 
    \item \textbf{RQ5}: Can we empirically verify $\ourSystem$'s scalability?
\end{enumerate}

\begin{table}[!tp]
\centering
\caption{The basic statistics of the datasets used in this paper.}
\vspace{-2.5mm}
\resizebox{\linewidth}{!}{
\begin{tabular}{l*{5}{c}r}
\hline
\bf Dataset &\bf  Amazon &\bf DBLP &\bf  Freebase &\bf  Twitter &\bf Healthcare\\
\hline
\# Nodes  & 10,099 &37,791 &14,541 & 9,990 &4,683\\
\# Edges   & 129,811 & 170,794 & 248,611  & 294,330 & 205,428\\
\# Relations & 2 &3 &237 & 4 &4\\
\# Training (positive) &126,535 &119,554 &272,115  & 282,115 &164,816 \\
\# Development &14,756 &51,242 &35,070 &32,926 &40,612 \\
\# Testing &29,492 &51,238 &40,932 &65,838 &40,612 \\\hline
\end{tabular}
}
\label{tab:datasets}
\end{table}

\subsection{Experimental Settings}
\subsubsection{Datasets used}
We use four public benchmark datasets covering multiple applications: \textit{e-commerce} (Amazon), \textit{academic graph} (DBLP), \textit{knowledge graphs} (Freebase), and \textit{social networks} (Twitter). We use the same data split for training, development, and testing as described in previous works~\cite{cen2019representation,abu2018watch,vashishth2019composition}. 
In addition, we also introduce a new knowledge graph from the publicly available real-world Medical Information Mart for Intensive Care (MIMIC) III dataset\footnote{https://mimic.physionet.org/} in the healthcare domain.
We generated equal number of positive and negative edges for the link prediction task. Table \ref{tab:datasets} provides the basic statistics of all datasets. The details of each dataset is provided below.

\begin{itemize}[leftmargin=*]
    \item \textbf{Amazon\footnote{https://github.com/THUDM/GATNE/tree/master/data \label{note2}}:} includes the co-viewing and co-purchasing links between products. The edge types, also\_bought and also\_viewed, represent that products are co-bought or co-viewed by the same user, respectively. 
    \item \textbf{DBLP\footnote{https://github.com/Jhy1993/HAN/tree/master/data}:} includes the relationships between papers, authors, venues, and terms. The edge types include paper\_has\_term, published\_at and has\_author. 
    \item \textbf{Freebase\footnote{https://github.com/malllabiisc/CompGCN/tree/master/data\_compressed}:} is a pruned version of FB15K with inverse relations removed. It includes links between people and their nationality, gender, profession, institution, place of birth/death, and other demographic features. 
    \item \textbf{Twitter}\textsuperscript{\ref{note2}}: includes links between tweets users. The edge types included in the network are re-tweet, reply, mention, and friendship/follower.
    \item \textbf{Healthcare:} includes relations between patients and their diagnosed medical conditions during each hospital admission along with relations to procedures and medications received. To ensure data quality, we use a 5-core setting, \textit{i.e.}, retaining nodes with at least five neighbors in the knowledge graph. The codes for generating this healthcare knowledge graph from MIMIC III dataset are also available at\footnote{https://github.com/pnnl/SLICE}.
\end{itemize}

\begin{table*}[!tp]
\centering
\caption{Performance comparison of different models on link prediction task using micro-F1 score and AUCROC. The symbol ``OOM'' indicates out of memory. Here, $\ourSystem_{w/o\ GF}$ and $\ourSystem_{w/o\ FT}$ represent two variants of the proposed $\ourSystem$ method by removing the Global Feature (GF) initialization and without fine-tuning (FT), respectively. The symbol * indicates that the improvement is statistically significant over the best baseline based on two-sided $t$-test with $p$-value $10^{-10}$.}
\vspace{-3.5mm}
\resizebox{1.0\linewidth}{!}{
\begin{tabular}{c|l|ccccc|ccccc}\hline
{\multirow{2}{*}{\textbf{Type}}}&{\multirow{2}{*}{\textbf{Methods}}}&\multicolumn{5}{c|}{\textbf{micro-F1 Score}} &\multicolumn{5}{c}{\textbf{AUCROC}} \\\cline{3-12}
& & \textbf{Amazon} & \textbf{DBLP} & \textbf{Freebase} & \textbf{Twitter} & \textbf{Healthcare} & \textbf{Amazon} & \textbf{DBLP} & \textbf{Freebase} & \textbf{Twitter} & \textbf{Healthcare}\\ \hline

{\multirow{4}{*}{\rotatebox[origin=c]{90}{\textbf{Static}}}}&\textbf{TransE} &50.28 &49.60  &47.78   &50.60 &48.42 &50.53 &49.05 &48.18 &50.26 &49.80\\
&\textbf{RefE} &51.86 &49.60  &50.25  &48.55 &47.96 &51.74 &48.50 &50.41 &49.28 &50.73\\

&\textbf{node2vec} &88.06 &86.71  &83.69   &72.72 &71.92 &94.48 &93.87  & 89.77  &80.48 &79.42\\

&\textbf{metapath2vec} &88.86 &44.58  &77.18    &66.73 &62.64   &95.42 &38.41  &84.33  &72.16 &69.11 \\\hline 

{\multirow{9}{*}{\rotatebox[origin=c]{90}{\textbf{Contextual}}}}&\textbf{GAN} &85.47 &OOM  &OOM   &85.01 &81.94 & 92.86 &OOM &OOM  &  92.39 &89.72 \\ 

&\textbf{GATNE-T} &89.06 &57.04  &OOM   &68.16 &58.02  &94.74 &58.44  &OOM     &72.07 &73.40\\

&\textbf{RGCN}  & 65.03 & 28.84 & OOM & 63.46 &56.73 & 74.77 & 50.35 & OOM & 64.35 &46.15\\

&\textbf{CompGCN} &83.42 &40.10  &65.39   &40.75 &39.84  &90.14 &34.04  &72.01  &39.86  &38.03 \\ 

&\textbf{HGT}  & 65.77 & 53.32 & OOM & 53.13 &76.54 & 68.66 & 50.85 & OOM & 59.32 &82.36\\

&\textbf{asp2vec} & \underline{94.89} & 78.82 & \underline{90.02} & \underline{88.29} & \underline{85.46} & \underline{98.51} & 92.51 &\bf 96.61 & \underline{95.00} & \underline{92.97}\\

&\textbf{$\ourSystem$}$_{w/o\ GF}$ &67.01 &66.02 &66.31 &67.07 &60.88  &62.87 &57.52 &55.31 &66.69 &63.11  \\
&\textbf{$\ourSystem$}$_{w/o\ FT}$ &94.99 &\underline{89.34} &90.01 &82.19 &81.58  &98.66 &\underline{96.07} &96.33 &90.38 &89.51  \\
&\textbf{$\ourSystem$} (Ours) &\bf 96.00* &\bf 90.70*  & \bf 90.26  &\bf 89.30* &\bf 91.64* &\bf 99.02* & \bf 96.69*  & \underline{96.41}   &\bf 95.73* &\bf 94.94* \\\hline

\end{tabular}
}
\vspace{-2mm}
\label{tab:f1-score}
\end{table*}

\subsubsection{Comparison Methods}
We compare our model against the following state-of-the-art network embedding learning methods. 
The first four methods learn static embeddings and the remaining methods learn contextual embeddings.

\begin{itemize}[leftmargin=*]
\item \textbf{TransE}~\cite{bordes2013translating} treats the relations between nodes as the translation operations in a low-dimensional embedding space.
\item \textbf{RefE}~\cite{chami2020low} incorporates hyperbolic space and attention-based geometric transformations to learn the logical patterns of networks. 
\item  \textbf{node2vec}~\cite{grover2016node2vec} is a random-walk based method that was developed for homogeneous networks. 
\item \textbf{metapath2vec}~\cite{dong2017metapath2vec} is an extension of node2vec that constrains random walks to specified metapaths in heterogeneous network.

\item \textbf{GAN} \textit{(Graph Attention Networks)}~\cite{abu2018watch} is a graph attention network for learning node embeddings based on the attention distribution over the graph walk context.  
\item \textbf{GATNE-T} (\textit{General Attributed Multiplex HeTerogeneous Network Embedding})~\cite{cen2019representation} is a metapath-constrained random-walk based method that learns relation-specific embeddings. 
\item \textbf{RGCN} \textit{(Relational Graph Convolutional Networks)} ~\cite{schlichtkrull2018modeling} learns multi-relational data characteristics by assigning a different weight for each relation. 
\item \textbf{CompGCN} \textit{(Composition-based Graph Convolutional Networks)} \cite{vashishth2019composition} jointly learns the embedding of nodes and relations for heterogeneous graph and updates a node representation with multiple composition functions. 
\item \textbf{HGT}\textit{ (Heterogeneous Graph Transformer)} ~\cite{hu2020heterogeneous} models the heterogeneity of graph by analyzing heterogeneous attention over each edge and learning dedicated embeddings for different types of edges and nodes. We adapt the released implementation of node classification task to perform link prediction task.
\item \textbf{asp2vec} \textit{(Multi-aspect network embedding)} ~\cite{park2020unsupervised} captures the interactions of the pre-defined multiple aspects with aspect regularization and dynamically assigns a single aspect for each node based on the specific local context.
\end{itemize}

\subsubsection{Experimental Settings} 
{All evaluations were performed using NVIDIA Tesla P100 GPUs. The results of $\ourSystem$ are evaluated under the} parameter settings described in Section \ref{sec: slice-implementation}. The results of all baselines are obtained with their original implementations. Note that for all baseline methods, the parameters not specially specified here are under the default settings.
We use the implementation provided in KGEmb\footnote{https://github.com/HazyResearch/KGEmb} for both \textbf{TransE} and \textbf{RefE}. \textbf{node2vec}\footnote{https://github.com/aditya-grover/node2vec} is implemented by sampling 10 random walks with a length of 80. The original implementation of \textbf{metapath2vec}\footnote{0https://ericdongyx.github.io/metapath2vec/m2v.html} is used by generating 10 walks for each node as well.  
We set the learning rate to be \{0.1, 0.01, 0.001\} and reported
the best performance for \textbf{GAN}\footnote{https://github.com/google-research/google-research/tree/master/\\graph\_embedding/watch\_your\_step}.
\textbf{GATNE-T} is implemented by generating 20 walks with a length of 10 for each node.
The results of \textbf{CompGCN}\footnote{https://github.com/malllabiisc/CompGCN} are obtained using the multiplicative composition of node and relation embeddings. 
We adapt the Deep Graph Library (DGL) based implementation\footnote{https://github.com/dmlc/dgl/tree/master/examples/pytorch/hgt} of \textbf{HGT} and \textbf{RGCN} to perform the link prediction task. 
We use the original implementation of \textbf{asp2vec}\footnote{https://github.com/pcy1302/asp2vec} for the evaluation. For all baselines, the dimension of embedding is set to 128.

\subsection{Evaluation on Link Prediction (RQ1)}
\label{ssec:link_prediction}
We evaluate the impact of contextual embeddings using the binary link prediction task, which has been widely used to study the structure-preserving properties of node embeddings~\cite{zhang2018link,chen2018pme}.

Table~\ref{tab:f1-score} provides the link prediction results of different methods on five datasets using micro-F1 score and AUCROC. The prediction scores for $\ourSystem$ are reported  from the context subgraph generation strategy (shortest path or random) that produces the best score for each dataset on the validation set. 
Compared to the state-of-the-art methods, we observe that $\ourSystem$ significantly outperforms both static and contextual embedding learning methods by 11.95\% and 25.57\% on average in F1-score, respectively. We attribute static methods superior performance, compared to relation based contextual learning methods (such as GATNE-T, RGCN, and CompGCN), to the ability of capturing global network connectivity patterns. Relation based contextual learning methods limit node contextualization by emphasizing the impact of relations on nodes. We outperform all methods on F1-score, including asp2vec, a cluster-aspect based contextualization method. Asp2vec achieves a marginally better AUCROC score on Freebase, but $\ourSystem$ achieves a better F1-score.

$\ourSystem$ outperforms asp2vec on all other datasets (in F1-score and AUCROC measure), improving F1-score for DBLP by 13\% owing to its ability to learn important metapaths without explicit specification.
These results indicate that subgraph based contextualization is highly effective and is a strong candidate for advancing the state-of-the-art for link prediction in a graph network.

\subsection{$\ourSystem$ Model Interpretation (RQ2)}

Here we study the impact of using different subgraph contexts on link prediction performance and demonstrate  $\ourSystem$'s ability to learn important higher-order relations in the graph. Our analysis shows $\ourSystem$' results are highly interpretable, and provide a way to perform explainable link prediction by learning the relevance of different context subgraphs connecting the query node pair.

\begin{figure}[!h]
	\centering
	\begin{subfigure}[b]{0.5\textwidth}
		\includegraphics[width=\textwidth]{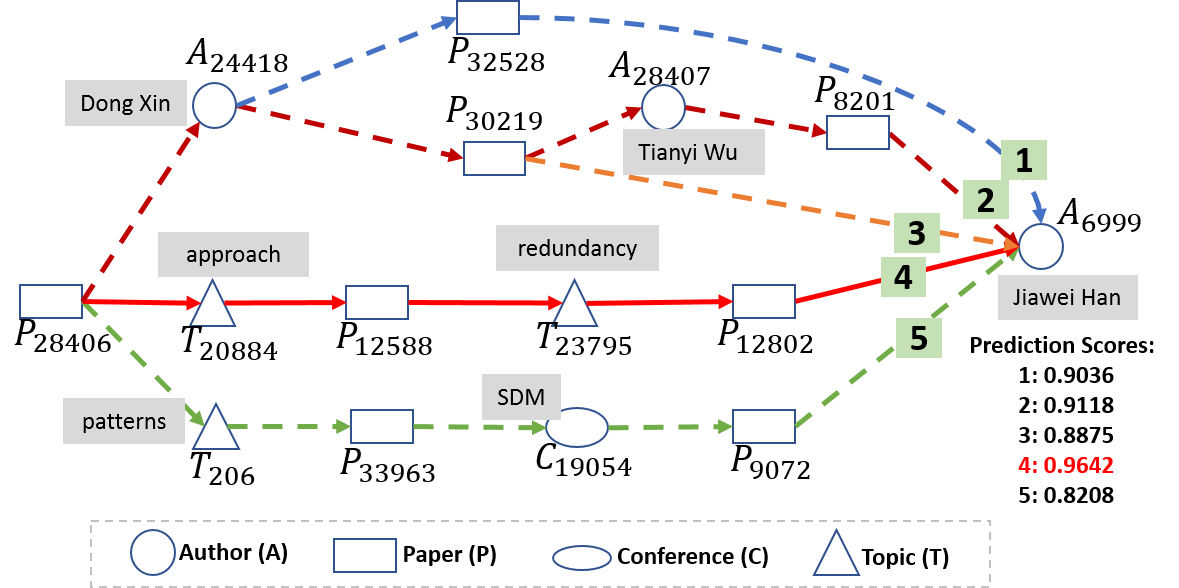}
		\vspace{-1mm}
	\end{subfigure}
	\begin{subfigure}[b]{0.5\textwidth}
    	\vspace{-2mm}
		\includegraphics[width=\textwidth]{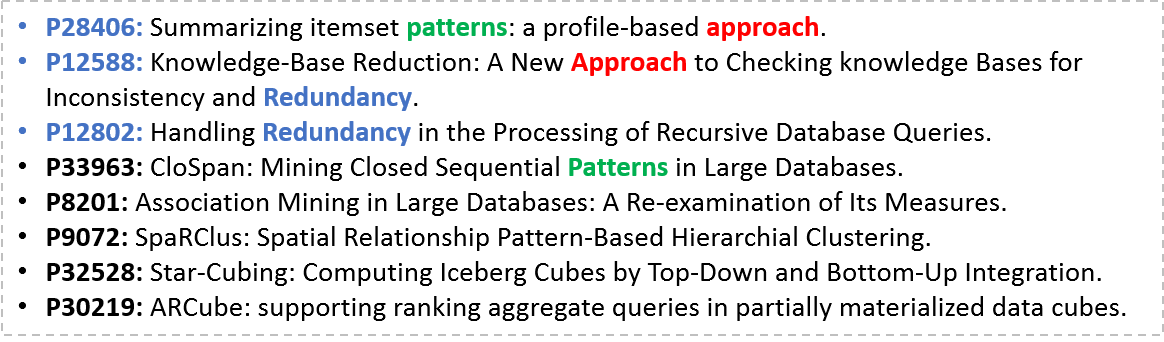}
	\end{subfigure}
	\vspace{-6mm}
	\caption{An example from DBLP. Relations in DBLP include paper-author, paper-topic and paper-conference. To predict the paper-author relationship between $P_{28406}$ and $A_{6999}$, five context subgraphs are generated with a beam-search strategy. The $4th$ context subgraph (along with context subgraph $1$, $2$ and $3$) contain closely related nodes and get high scores, while path $5$ containing a generic conference node achieves lowest score.}
	\label{fig:case-study}
	\vspace{-5mm}
\end{figure}

\begin{figure*}[!tp]
	\centering
	\begin{subfigure}[b]{0.225\textwidth}
		\includegraphics[width=\textwidth]{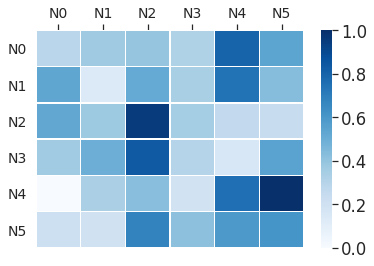}
		\vspace{-7mm}
		\caption{Layer 1}
		\label{fig:attention_visualization_a}
	\end{subfigure}
	\begin{subfigure}[b]{0.225\textwidth}
		\includegraphics[width=\textwidth]{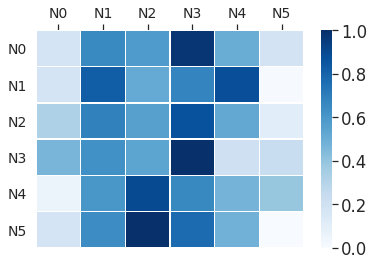}
		\vspace{-7mm}
		\caption{Layer 2}
		\label{fig:attention_visualization_b}
	\end{subfigure}
	\begin{subfigure}[b]{0.225\textwidth}
		\includegraphics[width=\textwidth]{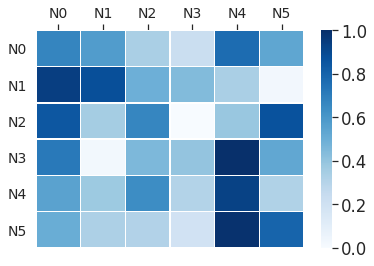}
		\vspace{-7mm}
		\caption{Layer 3}
		\label{fig:attention_visualization_c}
	\end{subfigure}
	\begin{subfigure}[b]{0.225\textwidth}
		\includegraphics[width=\textwidth]{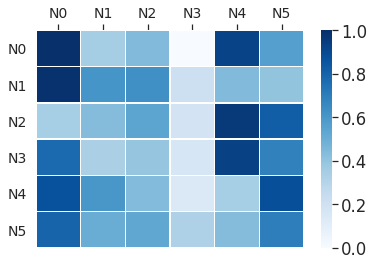}
		\vspace{-7mm}
		\caption{Layer 4}
		\label{fig:attention_visualization_d}
	\end{subfigure}
	\vspace{-3mm}
	\caption{Visualization of the semantic association matrix (after normalization) learnt from different layers on a DBLP subgraph for link prediction between paper N0 and author N1. An intense color indicates a higher association. Initially (layer 1), nodes N0 and N1 have low association. In layer 4, $\ourSystem$ learns higher semantic association from N1 to N0.}
	\vspace{-2mm}
	\label{fig:attention_visualization}
\end{figure*}

\subsubsection{\textbf{Case Study for Model Interpretation}}
Figure ~\ref{fig:case-study} shows an example subgraph from DBLP dataset between  ``Jiawei Han'' and ``$P_{28406}$'', a paper on frequent itemset mining. 
Paths 1 to 5 (shown in different legend) show different  contexts subgraphs present between the two nodes.  We observe that the link prediction score between Jiawei Han and paper $P_{28406}$ varies, depending on the context subgraph provided to the model.

We also observe that $\ourSystem$ assigns high link prediction scores for all context subgraphs (paths 1-4) where all the nodes on the path share strong semantic association with other nodes in the context subgraph.  The lowest score is achieved for path-5 which contains a conference node $C_{19054}$ (SIAM Data Mining). As a conference publishes papers in multiple topics, we hypothesize that this breaks the semantic association across nodes, and consequently lowers the probability of the link compared to other context subgraphs where all nodes are closely related.

It is important to note, that a node can share a semantic association with another node separated by multiple hops on the path, and thus would be associated via a higher-order relation.  We explore this concept further in the following subsections.

\subsubsection{\textbf{Interpretation of Semantic Association Matrix}}

We provide the visualization of the semantic association matrix $\bar{A}^{k}_{ij}$ as defined in Eq.~(\ref{eqn:embed_shift}) to investigate how the node dependencies evolve through different layers in $\ourSystem$.
Given a node pair ($v_i$, $v_j$) in the context subgraph $g_c$, a high value of $\bar{A}^{0}_{ij}$, indicates a strong global dependency of node $v_i$ on $v_j$. While a high value of $\bar{A}^{k}_{ij} (k\geq 1)$ (the association after applying more translation layers) indicates a prominent high-order relation in the subgraph context.

Figure~\ref{fig:attention_visualization} shows weights of semantic association matrix for the context generated for node pair (N0:~\textit{Summarizing itemset patterns: a profile-based approach (Paper)}, N1: \textit{Jiawei Han (Author)}). Nodes in the context consist of N2: \textit{Approach (Topic)}, N3: \textit{Knowledge-base reduction (Paper)}, N4:~\textit{Redundancy (Topic)} and N5: \textit{Handling Redundancy in the Processing of Recursive Database Queries (Paper)}. We observe that at layer 1 (Figure~\ref{fig:attention_visualization_a}), the association between source node \textit{N0} and target node \textit{N1} is relatively low. Instead, they both assign high weights on \textit{N4}. However, the dependencies between nodes are dynamically updated when applying more learning layers, consequently enabling us to identify higher-order relations. For example, the dependency of \textit{N1} on \textit{N0} becomes higher from layer 3~(Figure~\ref{fig:attention_visualization_c}) and \textit{N0} primarily depends on itself without highly influenced by other nodes in layer 4~(Figure~\ref{fig:attention_visualization_d}). This visualization of semantic association helps to understand how the global embedding is translated into the localized embedding for contextual learning. \vspace{0.04in}

\subsubsection{\textbf{Symbolic Interpretation of Semantic Associations via Metapaths}} 
Metapaths provide a symbolic interpretation of the higher-order relations in a heterogeneous graph.  We analyze the ability of $\ourSystem$ to learn relevant metapaths that characterize positive semantic associations in the graph.  We observe from Table \ref{tab:top_metapath} that $\ourSystem$ is able to match existing metapaths and also identify new metapath patterns for prediction of each relation type. For example, to predict the paper-author relationship,  $\ourSystem$ learns three shortest metapaths, including ``TPA" (authors publish with the same topic), ``APA" (co-authors) and ``CPA"(authors published in the same conference).

\begin{table}[!tp]
\centering
\caption{Comparisons of metapaths learned by $\ourSystem$ with predefined metapaths on DBLP dataset for each relation type. Here, P, A, C, and T represent Paper, Author, Conference, and Topic, respectively.}
\vspace{-3mm}
\resizebox{1.0\linewidth}{!}{
\begin{tabular}{l|ccc}
\hline
 \textbf{Learning Methods} & \textbf{Paper-Author} & \textbf{Paper-Conference} & \textbf{Paper-Topic} \\ \hline
Predefined \cite{yun2019graph}
 &APCPA, APA  &-  &-  \\
$\ourSystem$ + Shortest Path & TPA, APA, CPA  & TPC, APC, TPTPC  & TPT, CPT, APT  \\ 
$\ourSystem$ + Random & APA, APAPA  & TPTPC, TPAPC  & TPTPT, APTPT  \\\hline 
\end{tabular}
}
\vspace{-4mm}
\label{tab:top_metapath}
\end{table}

Interestingly, our learning suggests that longer metapath ``APCPA", which is commonly used to sample academic graphs for co-author relationship, is not as highly predictive of a positive relationship, i.e., ``all authors who publish in the same conference \emph{do not} necessarily publish together". Overall, the metapaths reported in Table \ref{tab:top_metapath} are consistent with the top ranked paths in Figure \ref{fig:case-study}.  These metapaths demonstrate $\ourSystem$'s ability to discover higher order semantic associations and perform interpretable link prediction in heterogeneous networks.

\begin{figure*}[htbp]
	\centering
	\begin{subfigure}[b]{0.19\textwidth}
		\includegraphics[width=\textwidth]{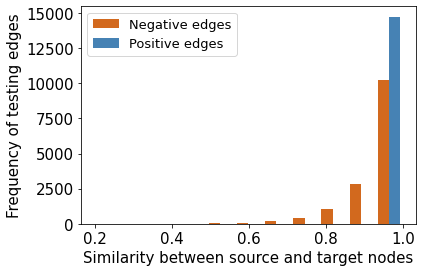}
		\caption{\small{node2vec: Amazon}}\label{fig:1a}
	\end{subfigure}
	\begin{subfigure}[b]{0.19\textwidth}
		\includegraphics[width=\textwidth]{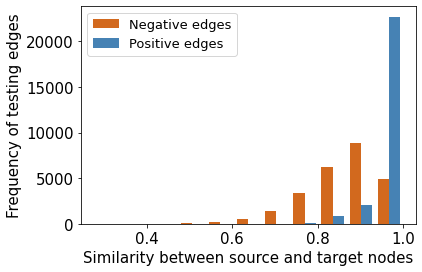}
		\caption{\small{node2vec: DBLP}}\label{fig:1a}
	\end{subfigure}
	\begin{subfigure}[b]{0.19\textwidth}
		\includegraphics[width=\textwidth]{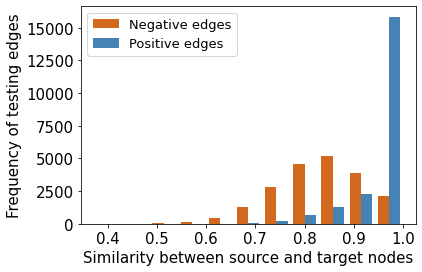}
		\caption{\small{node2vec: Freebase}}\label{fig:1a}
	\end{subfigure}
	\begin{subfigure}[b]{0.19\textwidth}
		\includegraphics[width=\textwidth]{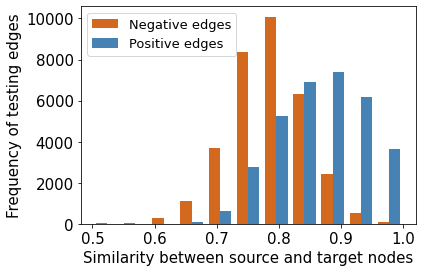}
		\caption{\small{node2vec: Twitter}}\label{fig:1c}
	\end{subfigure}
	\begin{subfigure}[b]{0.19\textwidth}
		\includegraphics[width=\textwidth]{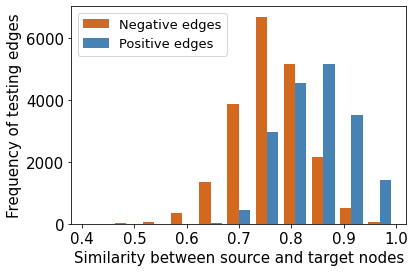}
		\caption{\small{node2vec: Healthcare}}\label{fig:1c}
	\end{subfigure}
	
	\begin{subfigure}[b]{0.19\textwidth}
		\includegraphics[width=\textwidth]{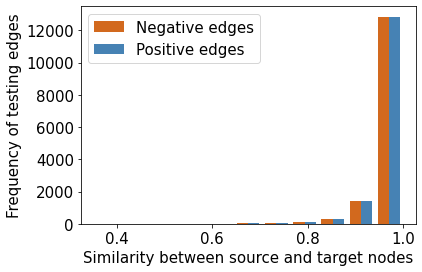}
		\caption{\small{CompGCN: Amazon}}\label{fig:1b}
	\end{subfigure}
	\begin{subfigure}[b]{0.19\textwidth}
		\includegraphics[width=\textwidth]{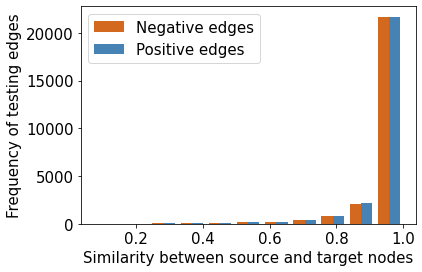}
		\caption{\small{CompGCN: DBLP}}\label{fig:1b}
	\end{subfigure}
	\begin{subfigure}[b]{0.19\textwidth}
		\includegraphics[width=\textwidth]{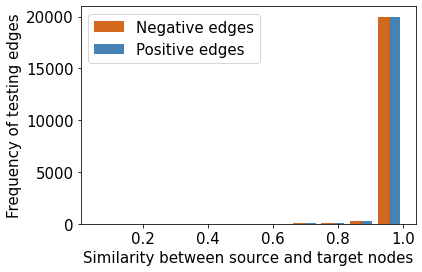}
		\caption{\small{CompGCN: Freebase}}\label{fig:1a}
	\end{subfigure}
	\begin{subfigure}[b]{0.19\textwidth}
		\includegraphics[width=\textwidth]{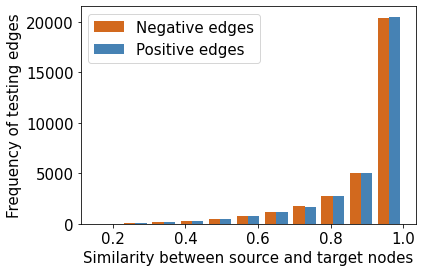}
		\caption{\small{CompGCN: Twitter}}\label{fig:1d}
	\end{subfigure}
	\begin{subfigure}[b]{0.19\textwidth}
		\includegraphics[width=\textwidth]{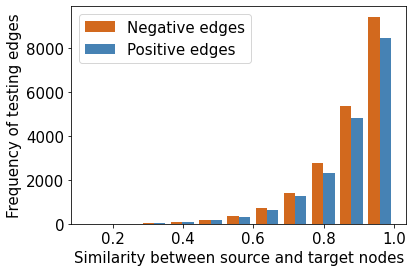}
		\caption{\small{CompGCN: Healthcare}}\label{fig:1d}
	\end{subfigure}
	
	\begin{subfigure}[b]{0.19\textwidth}
		\includegraphics[width=\textwidth]{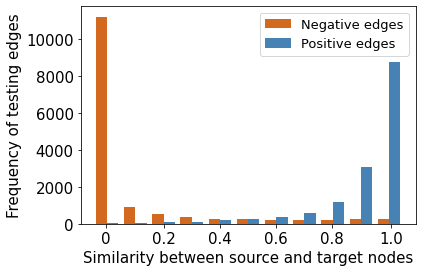}
		\caption{\small{$\ourSystem$: Amazon}}\label{fig:1b}
	\end{subfigure}
	\begin{subfigure}[b]{0.19\textwidth}
		\includegraphics[width=\textwidth]{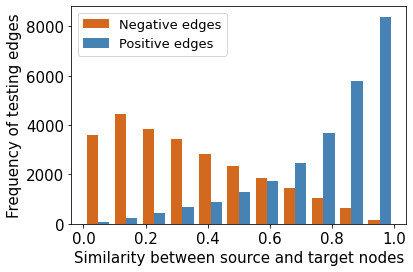}
		\caption{\small{$\ourSystem$: DBLP}}\label{fig:1b}
	\end{subfigure}
	\begin{subfigure}[b]{0.19\textwidth}
		\includegraphics[width=\textwidth]{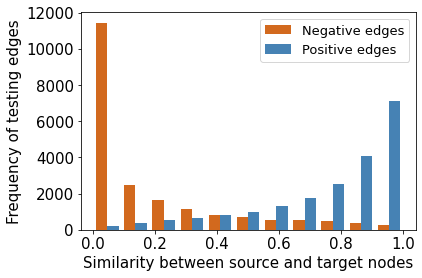}
		\caption{\small{$\ourSystem$: Freebase}}\label{fig:1a}
	\end{subfigure}
	\begin{subfigure}[b]{0.19\textwidth}
		\includegraphics[width=\textwidth]{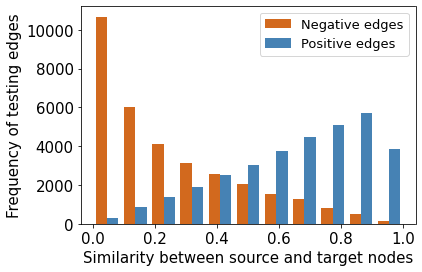}
		\caption{\small{$\ourSystem$: Twitter}}\label{fig:1d}
	\end{subfigure}
	\begin{subfigure}[b]{0.19\textwidth}
	    \includegraphics[width=\textwidth]{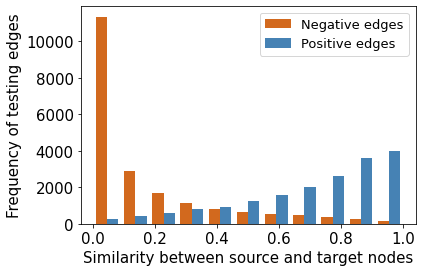}
		\caption{\small{$\ourSystem$: Healthcare}}\label{fig:1d}
	\end{subfigure}
	\vspace{-2mm}
	\caption{Distributions of similarity scores of both positive and negative node-pairs obtained by node2vec, CompGCN, and $\ourSystem$ over five datasets.}
	\label{fig:distribution_of_similarity_2}
\end{figure*}

\subsection{Effectiveness of Contextual Translation for Link Prediction (RQ3)}
In this section, we study the impact of contextual translation on node embeddings. First, we evaluate the impact of contextualization in terms of the similarity (or distance) between the query nodes. Second, we analyze the effectiveness of contextualization as a function of the query node pair properties. The latter is especially relevant for understanding the performance boundaries of the contextual methods.

\subsubsection{\textbf{Impact of Contextual Translation on Embedding-based Similarity}}

Figure \ref{fig:distribution_of_similarity_2} provides the distribution of similarity scores for both positive and negative edges obtained by $\ourSystem$. We compare against embeddings produced by node2vec~\cite{grover2016node2vec} which is one of the best performing static embedding methods (see Table \ref{tab:f1-score}) and CompGCN~\cite{vashishth2019composition} a relation based contextualization method. We observe that for node2vec and CompGCN, the distribution of similarity scores across positive and negative edges overlaps significantly for all datasets. This indicates that the embeddings learnt from global methods or relation specific contextualization cannot efficiently differentiate the positive and negative edges in link prediction task.

On the contrary, $\ourSystem$ increases the margin between the distributions of positive and negative edges significantly. It brings node embeddings in positive edges closer and shifts nodes in negative edges farther away in the low-dimensional space. This indicates that the generated subgraphs provide informative contexts during link prediction and enhance embeddings such that it improves the discriminative capability of both positive and negative node-pairs. 

\begin{figure}[!tp]
	\centering
	\begin{subfigure}[b]{0.236\textwidth}
		\includegraphics[width=\textwidth]{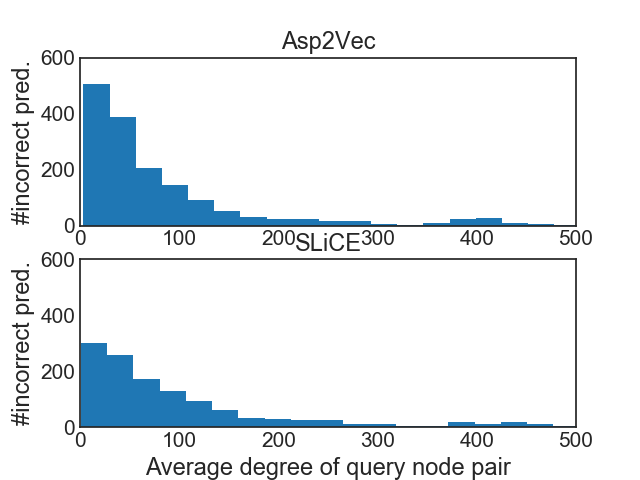}
		\caption{Amazon}
		\label{fig:asp2vec_comp_amazon}
	\end{subfigure}
	\begin{subfigure}[b]{0.236\textwidth}
		\includegraphics[width=\textwidth]{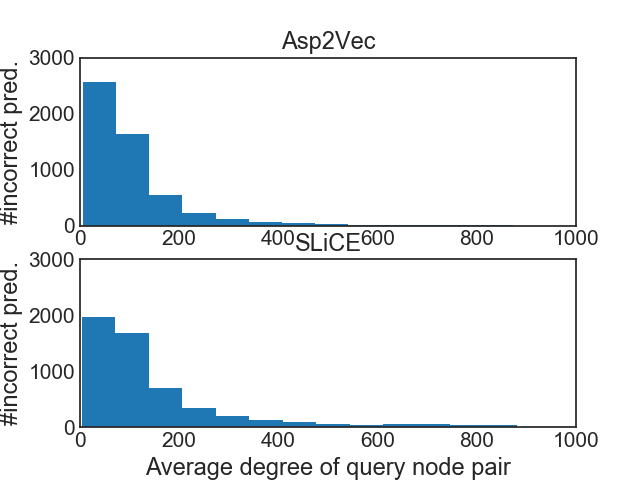}
		\caption{Healthcare}
		\label{fig:asp2vec_comp_mimic}
	\end{subfigure}
	\vspace{-6mm}
	\caption{Error analysis of contextual translation based link prediction method as a function of degree-based connectivity of query nodes.}
	\vspace{-2mm}
	\label{fig:asp2vec_comp_error_profile}
\end{figure}

\subsubsection{\textbf{Error Analysis of Contextual Methods as a function of Node Properties}}
We investigate the performance of $\ourSystem$ and the closest performing contextual learning method, asp2vec \cite{park2020unsupervised}, as a function of query node pair properties. Given each method, we select all query node pairs drawn from both positive and negative samples that are associated with an incorrect prediction. Next, we compute the average degree of the nodes in each such pair.  We opt for degree and ignore any type constraint for it's simplicity and ease of interpretation.  Fig. \ref{fig:asp2vec_comp_error_profile} shows the distribution of these incorrect predictions as a function of average degree of query node pair. It can be seen that, for Amazon and Healthcare datasets, most of the incorrect predictions are concentrated around the query pairs with low and medium values of average degree.  However, $\ourSystem$ has fewer errors than asp2vec for such node pairs. This can be attributed to the \textsl{aspect-oriented} nature of asp2vec, which maps each node to a fixed number of aspects. Since nodes in a graph may demonstrate varying degree of aspect-diversity, mapping a node with low diversity to more aspects (than it belongs to) reduces asp2vec's performance. $\ourSystem$ adopts a complementary approach, where it considers the subgraph context that connects the query nodes, leading to better contextual representations.

\begin{figure*}[htbp]
	\centering
	\begin{subfigure}[b]{0.235\textwidth}
		\includegraphics[width=\textwidth]{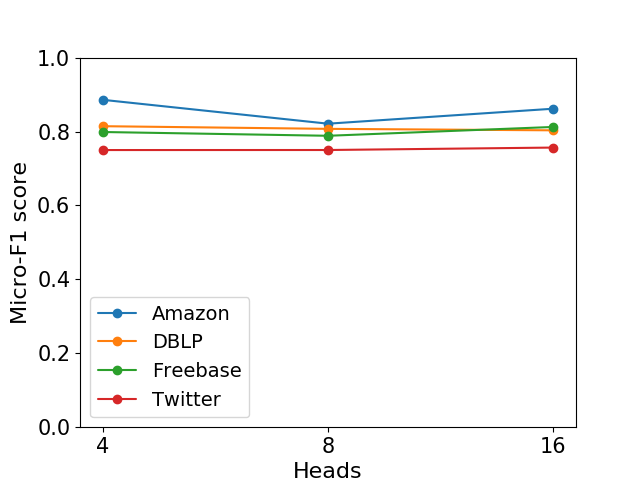}
		\caption{}\label{fig:1a}
	\end{subfigure}
	\begin{subfigure}[b]{0.235\textwidth}
		\includegraphics[width=\textwidth]{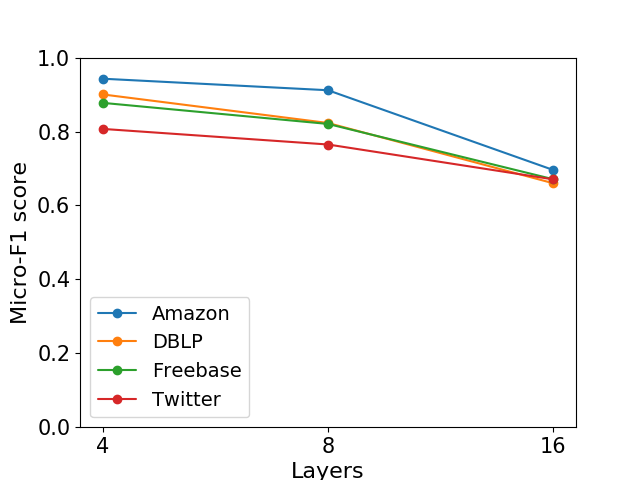}
		\caption{}\label{fig:1b}
	\end{subfigure}
	\begin{subfigure}[b]{0.235\textwidth}
		\includegraphics[width=\textwidth]{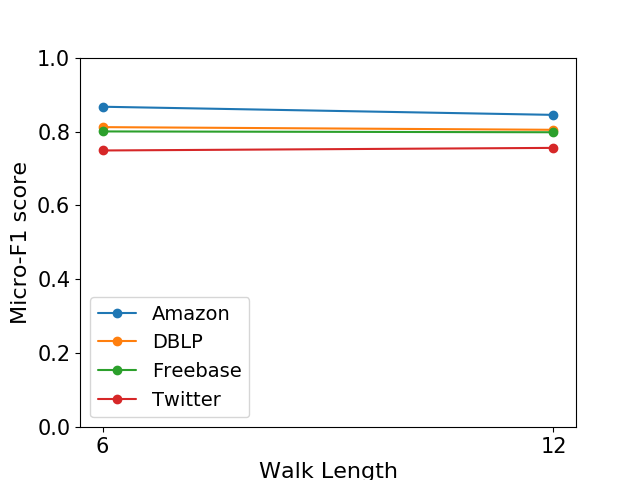}
		\caption{}\label{fig:1c}
	\end{subfigure}
	\begin{subfigure}[b]{0.235\textwidth}
		\includegraphics[width=\textwidth]{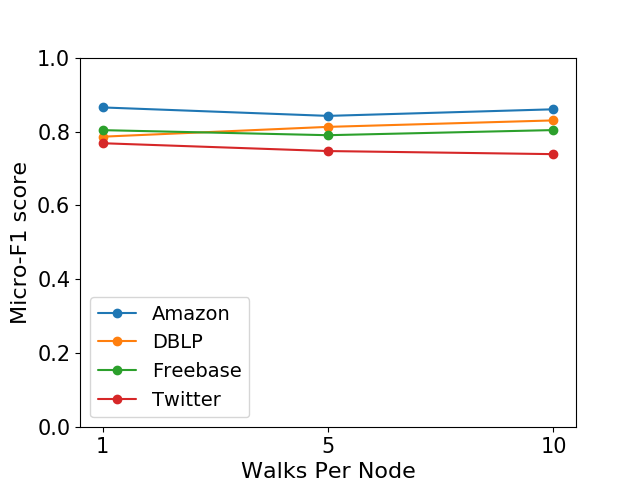}
		\caption{}\label{fig:1d}
	\end{subfigure}
	\vspace{-2mm}
	\caption{Micro-F1 scores for link prediction with different parameters in $\ourSystem$ on four datasets.}
	\label{fig:parameter_sensitivity_1}
\end{figure*}

\begin{figure}[!tp]
	\centering
	\begin{subfigure}[b]{0.235\textwidth}
		\includegraphics[width=\textwidth]{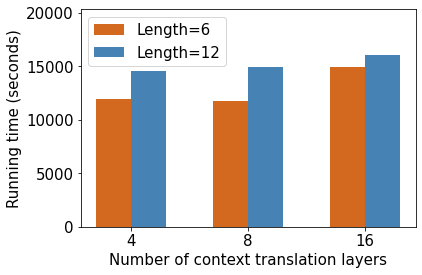}
		\caption{}
		\label{fig:complexity-1}
	\end{subfigure}
	\begin{subfigure}[b]{0.235\textwidth}
		\includegraphics[width=\textwidth]{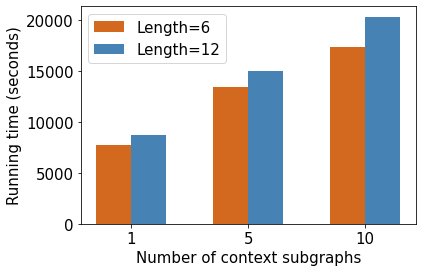}
		\caption{}
		\label{fig:complexity-2}
	\end{subfigure}
	\vspace{-4mm}
	\caption{Analysis of the time complexity of $\ourSystem$ on Freebase dataset by varying (a) number of translation layers and (b) number of context subgraphs considered for each node.}
	\vspace{-4mm}
	\label{fig:complexity}
\end{figure}

\subsection{Study of \ourSystem~(RQ4)}

\subsubsection{\textbf{Parameter Sensitivity}}
In Figure \ref{fig:parameter_sensitivity_1}, we provide the link prediction performance with micro-F1 score on four datasets by varying four parameters used in $\ourSystem$ model, including number of heads, number of contextual translation layers, number of nodes in contexts (\textit{i.e.}, walk length), and the number of (context) walks generated for each node in pre-training. The performance shown in these plots are the averaged performance by fixing one parameter and varying other three parameters. 

On the other hand, when varying the parameter \textit{number of layers}, we observe that applying four layers of contextual translation provides the best performance on all the datasets; the performance dropped significantly when stacking more layers. This indicates that four contextual translation layers are sufficient to capture the complex higher-order relations over various knowledge graphs. Based on these analysis, we set the default values for both number of heads and the number of layers to be 4, and generate one walk for each node with a length of 6 in the pre-training step.

\subsubsection{\textbf{Effect of Pre-trained Global Features (GF)}}
To explore the impact of pre-trained global node features on the performance of $\ourSystem$, we performed an ablation study by analyzing a variant of $\ourSystem$ with four contextual translation layers. More specifically, the pre-trained global embeddings are disabled in $\ourSystem$, termed $\ourSystem_{w/o\ GF}$. The results are provided in Table \ref{tab:f1-score}. 
We observe that without initialization using the pre-trained global embeddings, the model performance of $\ourSystem$ decreased on all five datasets in both metrics. The reason being that, compared to the random initialization, the global node features are able to represent the role of each node in the global structure of the knowledge graph. By further applying the proposed contextual translation layers, they can collaboratively and efficiently provide contextualized node embeddings for downstream tasks like link prediction.

\subsubsection{\textbf{Effect of Fine-tuning (FT)}} 
To investigate the effect of fine-tuning stage on learning the contextual node embeddings, we disable the fine tuning layer for supervised link prediction task, termed $\ourSystem_{w/o\ FT}$ and show the results in Table~\ref{tab:f1-score}. Compared to the baseline methods, it  still achieves competitive performance. We attribute this to the effectiveness of capturing higher-order relations through the contextual translation layers. While compared to the full $\ourSystem$ model, the performance of $\ourSystem_{w/o\ FT}$ degrades slightly on Amazon, DBLP, and Freebase datasets, but significantly decreases on both Twitter and MIMIC datasets. This can be attributed to the fact that supervised training with the link prediction task is able to learn the fine-grained contextual node embedding for link prediction task.

\begin{table}[!tp]
\centering
\caption{Estimation of the number of context subgraphs for each node in the knowledge graph.}
\vspace{-2mm}
\resizebox{1.0\linewidth}{!}{
\begin{tabular}{l*{5}{c}r}
\hline
\bf Dataset &\bf  Amazon &\bf DBLP &\bf  Freebase &\bf  Twitter &\bf Healthcare\\
\hline
\# Nodes ($|V|$)  & 10,099 &37,791 &14,541 & 9,990 &4,683\\
\# Edges ($|E|$)   & 129,811 & 170,794 & 248,611 &294,330 &205,428 \\
\# Contexts ($N_{cpn}$) &7.74 &2.71 &10.26 &17.67 &26.32 \\\hline
\end{tabular}
}
\label{tab:sample_analysis}
\vspace{-2mm}
\end{table}
\subsection{$\ourSystem$ Model Complexity (RQ5)}

Contextual learning methods are known to have high computational complexity. In section \ref{sec:complexity} we observed that the cost of training $\ourSystem$ is approximately linear to the number of edges. In this section, we provide an empirical evaluation of the model scalability in view of the prior analysis.

\subsubsection{\textbf{Time Complexity Analysis}}
In this subsection, we mainly investigate the impact of the following three parameters on the overall time complexity of the $\ourSystem$ model: (1) number of contextual translation layers, (2) number of context subgraphs, and (3) length of context subgraph.

\begin{itemize}[leftmargin=*]
    \item We study the scalability of the $\ourSystem$ model when the number of context translation layers are varied and the corresponding plots are provided in Figure \ref{fig:complexity}(a). The $x$-axis and $y$-axis represent the number of layers and running time in seconds, respectively. The plots indicate that increasing the number of layers does not significantly increase the training time of the $\ourSystem$ model. 
    \item In Figure \ref{fig:complexity}(b), we demonstrate the impact of the number of context subgraphs on the time complexity. Increasing the number of context subgraphs generated for each node in pre-training and each node-pair for fine-tuning raises the number of training edges which further increases the training time of the model.  
\end{itemize}
These two plots empirically verify the analysis about the model complexity, discussed in Section \ref{sec:complexity}, that the proposed $\ourSystem$ model is approximately linear to the number of edges in the graph and does not depend on other parameters such as the number of contextual translation layers and the number of nodes in the graph. In addition, we also vary the length (number of nodes) of the context subgraph in both plots. The plot shows that even doubling the context length will not significantly increase the running time.  This time complexity analysis, combined with the performance results in Table \ref{tab:f1-score} and the parameter sensitivity analysis in Figure \ref{fig:parameter_sensitivity_1} can jointly provide the guidelines for parameter selection.

\subsubsection{\textbf{Context Subgraph Sampling Analysis}}
In the complexity analysis discussed in Section \ref{sec:complexity}, we approximated the total number of training edges in the entire graph as $N_E \approx |V|*N_{cpn}$, where $|V|$ represents the number of nodes in graph $G$ and $N_{cpn}$ denotes the number of context subgraphs generated for each node. This estimation also provides us guidelines for determining the number of context subgraphs for each node $N_{cpn}$. By incorporating $N_E = \alpha_T |E|$ into the approximation (where $|E|$ is the total number of edges in graph $G$ and $\alpha_T$ is the ratio of training split), we can estimate the number of context subgraphs per node as $N_{cpn}=\alpha_T |E|/|V|$. Table~\ref{tab:sample_analysis} shows the estimated numbers (with $\alpha_T=0.6$) for the five datasets used in this work. These estimations provide us an approximate range for the value of $N_{cpn}$ during the context generation step. Based on this analysis, in our experiments, we generally consider 1, 5, and 10 for the value of $N_{cpn}$ on all the five datasets in both pre-training and fine-tuning stages.

\section{Conclusions}
\label{sec:conclusion}
We introduce $\ourSystem$ framework for learning contextual subgraph representations. Our model brings together knowledge of structural information from the entire graph and then learns deep representations of higher-order relations in arbitrary context subgraphs. $\ourSystem$ learns the composition of different metapaths that characterize the context for a specific task in a drastically different manner compared to existing methods which primarily aggregate information from either direct neighbors or semantic neighbors connected via certain pre-defined metapaths. $\ourSystem$ significantly outperforms several competitive baseline methods on various benchmark datasets for the link prediction task. In addition to demonstrating $\ourSystem$'s interpretability and scalability, we provide a thorough analysis on the effect of contextual translation for node representations. In summary, we show $\ourSystem$'s subgraph-based contextualization  approach is effective and distinctive over competing methods.

\begin{acks}
This work was supported in part by the US National Science Foundation grant IIS-1838730, Amazon AWS cloud computing credits, and Pacific Northwest National Laboratory under DOE-VA-21831018920.
\end{acks}

\bibliographystyle{ACM-Reference-Format}
\balance
\bibliography{citations}


\end{document}